\documentclass[preprint,11pt,authoryear]{elsarticle}

\usepackage{natbib}
\usepackage[margin=1in]{geometry}
\usepackage{todonotes}

\usepackage{amssymb}

\usepackage{booktabs}
\usepackage{float}
\usepackage{xurl}
\usepackage{yfonts}

\journal{tba}

\usepackage[hidelinks]{hyperref}

\begin{document}

\begin{frontmatter}



\title{Basic syntax from speech: Spontaneous concatenation in unsupervised deep neural networks}

\author[label1]{Ga\v{s}per Begu\v{s}\corref{cor1}}
\ead{begus@berkeley.edu}
\author[label1]{Thomas Lu}
\author[label2]{Zili Wang}

\affiliation[label1]{organization={Department of Linguistics, University of California, Berkeley},
            country={United States}}

            \affiliation[label2]{organization={Department of Computer Science, Iowa State University},
            country={United States}}

\begin{abstract}

Computational models of syntax are predominantly text-based. Here we propose that the most basic first step in the evolution of syntax can be modeled directly from raw speech in a fully unsupervised way. We focus on one of the most ubiquitous and elementary suboperations of syntax---concatenation. We introduce \textit{spontaneous concatenation}: a phenomenon where a ciwGAN/fiwGAN models (based on convolutional neural networks) trained on acoustic recordings of individual words start generating outputs with  two or even three words concatenated without ever accessing data with multiple words in the training data. We replicate this finding in several independently trained models with different hyperparameters and training data. Additionally, networks trained on two words learn to  embed words into novel unobserved word combinations. We also show that the concatenated outputs contain precursors to compositionality. To our knowledge, this is a previously unreported property of CNNs trained in the ciwGAN/fiwGAN setting on raw speech and has implications both for our understanding of how these architectures learn as well as for modeling syntax and its evolution in the brain from raw acoustic inputs. We also propose and formalize a neural mechanism called \textit{disinhibition} that outlines a possible artificial and biological neural pathway towards concatenation and compositionality and suggests our modeling is useful for generating testable predictions for biological and artificial neural processing of spoken language.

\end{abstract}




\end{frontmatter}



\section{Introduction}

How language evolved and what is required for humans to acquire syntax are among the central questions in linguistics and cognitive science \citep{christiansen03,nowak01,fitch00}. Due to lack of direct evidence for language evolution, many studies employ evolutionary modeling \citep{christiansen03a,ogrady18} in order to gain insights into the putatively most difficult problem in science. While many studies achieved substantial progress \citep{nowak99,kirby00,griffiths07,zuidema09,perfors14,ogrady18}, most models, however, are either theoretical or operate with highly simplified and abstract assumptions. To our knowledge, concatenation does not emerge spontaneously in any of the existing models. In this paper, we model the first step in the evolution/acquisition of syntax---concatenation---with unsupervised deep neural networks that are among the most realistic models of human language learning and learn linguistic representations from raw spoken language inputs \citep{begusCiw}. 

Concatenation (or compounding/conjoining elements) is one of the most basic operations in human language.  Many (but not all) animal communication systems   use simple symbols (call/sign$\sim$meaning pairs) that are not concatenated. Such  holistic calls have been termed ``elementary signals'' by \citet{nowak01}. In human language, on the other hand, individual elements such as words can combine into ``compound signals'' \citep{nowak01} with compositional meaning. The evolution of concatenation \citep{jackendoff99,luuk14,progovac15} as well as the existence of related operations that are presumably uniquely human and domain-specific have been the focus of debates in linguistics and cognitive science. In Jackendoff's (\citeyear{jackendoff99}) model, concatenation is the first step in the evolution of syntax. \citet{luuk14} propose a stage in evolution of syntax with concatenated symbols that are not yet fully compositional. They back their model with empirical evidence from language acquisition, where it has been shown that children concatenate words without a more complex embedding operation \citep{diessel05}. Concatenation plays a central role in  theoretical approaches to human syntax as well. One of the influential syntactic frameworks, the minimalist program  \citep{chomsky95}, introduces an operation called \textit{Merge}, which in addition to other suboperations contains concatenation. Concatenation, or the switch from a holistic single unit system towards multiple-unit compound signals, is the first and crucial step in both evolution and acquisition of human syntax. It remains unclear whether domain-specific mechanisms (such as Merge) are required for the development towards concatenated signals or whether concatenation can arise spontaneously in domain-general learning models. This paper aims to answer this question.

Concatenation in itself is not yet syntax.   Many other systems, such as bird song \citep{berwick12,okanoya13,youngblood20} or music \citep{patel07} contain concatenation with no referential meaning.  The degree to which these systems are analyzed in terms of syntactic processing is still disputed.  In our proposal,  meaning is modeled in a highly abstract way with binary codes.   This is only an imperfect, but not insurmountable approximation of the actual referential meaning. Concatenating raw phonetic items into multiple-word outputs is thus the very first and most basic precursor to syntax, and not syntax itself. However, without this step, no further syntactic evolution could have been possible. Human syntax is compositional. We show that our model not only spontaneously concatenates, but also does so in a compositional manner for at least a subset of lexical items, which presents an additional step towards the evolution of human syntax.

In this paper, we propose the following terminology. Outputs that contain a single lexical item based on a learned symbolic-like representation of that item are analyzed as simple signals. When an output contains two  simple signals that can otherwise independently surface in the output, we call this a concatenated signal output. Because our models clearly distinguish one output from another (one generation from the next generation), the presence of two items in a single output is a clear sign of concatenation. When such concatenation occurs in a model that had no concatenated signals in the training data, we call such concatenation \textit{spontaneous}. When the learned symbolic-like representations of each individual simple signal causally and predictably represent the  concatenation of those items for  at least a subset of concatenated outputs, we call this a semi-compositional concatenation. When such representation works for all items predictably, such a system would be a fully compositional concatenation. Compositional concatenation is a precursor to human syntax and proposed operations such as \textit{Merge} \citep{chomsky95}. Merge is more complex than compositional concatenation, because it can operate recursively on already merged items. These levels of complexity are useful for comparing the models' performance with human language, animal communications, and language acquisition stages. For example, it has been argued that language-acquiring children undergo a development from a stage of simple signals to  non-compositional concatenation to compositional syntax \citep{diessel05,luuk14}. Various degrees of compositional concatenation have been observed in animal communication \citep{outtara09,suzuki16}.

As our model of learning we adopt ciwGAN and fiwGAN \citep{begusCiw} which model language as informative imitation, feature several advantages over competing computational models, and represent one of the closest approximations of human language learning using deep neural networks. The models need to learn to generate spoken language from some low-dimension random representations by only listening to raw audio. The learning happens by informative imitation: the Generator needs to learn to generate data such that the Discriminator will be inefficient in distinguishing between the internally generated data and external training data. In human processing terms, the model needs to learn to build internal representations such that the generated data approximates what another network hears as primary linguistic data. This constitutes the imitation principle in fiwGAN/ciwGAN learning. Crucially, the models also need to exchange information through the imitated vocalizations. The Generator needs to learn to transform binary or one-hot codes into spoken language, which the Q-network decodes. 

The model has the generative component mimicking human speech production (the Generator network), the classification component mimicking human speech perception (the Q-network) \citep{begus22interspeech},  and the adversarial component mimicking the imitation principle in human speech acquisition. The models have recently been shown to closely mimic early auditory processing in human brain based on language experience \citep{begusZhouZhao23}. The models have also been shown to closely follow stages in spoken language acquisition \citep{begus19} and can be directly compared to the behavioral experiments (e.g.~in vowel harmony learning; \citealt{begusLocal}). The production-perception loop is the condition for learning in these models. There are important limitations to the models as well. The meaning in the models is currently represented in information theoretic terms with either binary codes or one-hot vectors and not with grounded visual information. For the purpose of this paper, we use the ciwGAN/fiwGAN family of models that do not include articulators; however, we expect similar results in the ciwaGAN \citep{begusCiwa} model where the Generator produces articulatory movements rather than raw audio. 

The sounds of human speech are a measurable, physical property, but they also encode abstract linguistic information such as syntactic, phonological, morphological, and semantic properties. It has been shown that the model learns symbolic-like rule-like linguistic representations from raw speech \citep{begusCiw,begus2020identity,begus19,begusLocal} which illustrates that discrete mental representational units can emerge from continuous physical properties such as sound  in deep neural networks. The model features several aspects of human spoken language learning that other models lack: learning by imitation, learning through the production-perception loop, and communicative intent.  

Prior models of human syntax are predominantly text-based and mostly lack explicit mechanisms that can be directly paralleled to language processing in the human brain. Hearing human learners, however, acquire syntax from acoustic inputs and process it with biological neural mechanisms. Few studies model syntax from speech with artificial neural networks. Previous work has shown that speech-trained models can feature syntactic representations, but the models featured specific syntactic mechanisms \citep{lai23} or were pre-trained \citep{singla22}. These architectures do not model the emergence of syntax spontaneously.

Here, we propose that we can model a shift from single unit system to a concatenative and partially compositional system using ciwGAN. Our models are trained on raw speech, which contains both the presence of a signal (a word) as well as the absence thereof (no word). This provides a good testing ground for modeling concatenative and compositional learning in general. We model how compound signals or concatenated words can arise spontaneously in deep neural networks in a fully unsupervised manner. We thus also provide evidence suggesting that domain-general learning models (such as CNNs) spontaneously concatenate signals in a learning setting that involves informative imitation: learning by imitation and with the production-perception loop from raw spoken language data (ciwGAN). The paper thus introduces a model of the first step in the development from holistic non-syntactic signals to compositionally concatenated signals that represents the precursor to  syntactic processing. We also show that the observed spontaneous concatenation is not an idiosyncratic property of a single model, but emerges in several replicated models as well as in models that feature several changes in the architecture.  

In Section \ref{disinhibition}, we provide a plausible neural explanation for the concatenative behavior called the \textit{disinhibition} which suggests that such modeling can be useful for generating and simulating predictions for neural processing of spoken language. Disinhibition is a well-documented neural process whereby neurons inhibit inhibitory neurons which results in disinhibition or excitation \citep{pi13}, but has not been connected to compositional concatenation yet.  We also show that the networks exhibit traces of compositionality (Section \ref{compositionality}), which represent an additional step toward the evolution of human syntax. In Section \ref{formalizingdisinhibition}, we formalize the interaction between latent variables and connect the artificial disinhibition to existing mathematical treatments of the Merge operation.

\section{Methods}

To test whether CNNs can spontaneously concatenate lexical items, we conduct two sets of experiments. In the first set of experiments (four \textit{one-word} experiments), we train the networks on single-word inputs and test the networks for concatenated outputs.  In the second experiment, we train the networks on one-word and two-word inputs (the \textit{two-word} experiment) and withhold a subset of two-word combinations. We then test whether words can be embedded into novel unobserved combinations in the output. 

Because the networks are trained in the GAN setting, the Generator never accesses the training data directly, but generates innovative outputs. It has been shown that GANs innovate in highly interpretable ways that produce novel words or sound sequences \citep{begusCiw}. Here we test whether innovations can produce spontaneously concatenated words.

\subsection{The model}
We train the ciwGAN and modified fiwGAN models \citep{begusCiw}. CiwGAN/fiwGAN models are information-theoretic  extensions of GANs (based on InfoGAN, WaveGAN and DCGAN; \citealt{chen16,donahue19,radford15}) designed to learn from audio inputs.

\begin{figure}
    \centering
    \includegraphics[width=.8\textwidth]{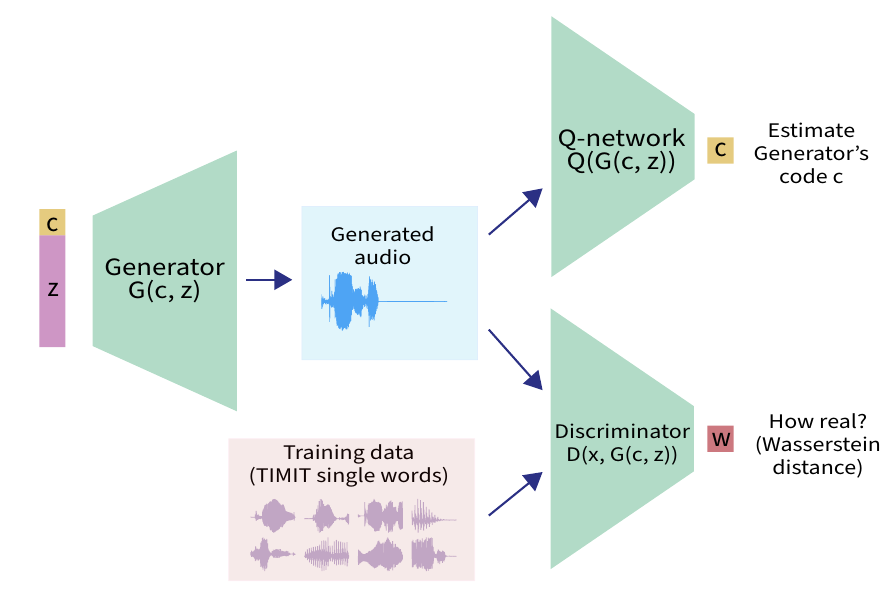}
    \caption{The architecture of ciwGAN used in the two-second one-word experiment.}
    \label{fig:StructureCiwgan}
\end{figure}

    The ciwGAN/fiwGAN architectures  involve three networks (Figure \ref{fig:StructureCiwgan} and Table \ref{structure}): the Generator (G) that takes latent codes $c$ (either one-hot vectors or binary codes) and random latent space variables $z$ ($z\sim\mathcal{U}(-1,1)$) and through five or six upconvolutional layers generates 1.024s or 2.048s audio with 16 kHz sampling rate (16,384 or 32,768 samples). The audio is then fed to the Discriminator (D), which evaluates realness of the output via the Wasserstein loss (WGAN) \citep{arjovsky17}. The model is formalized in \citet{begusCiw} as in (\ref{gans1}).

    \begin{equation}\label{gans1}
\min_{G,Q}\max_{D} V_{IWGAN}(D, G, Q) = V_{WGAN}(D, G) -\lambda L_I (G, Q).
\end{equation}
    
    The unique aspect of the ciwGAN/fiwGAN architecture is a separate Q-network, which is trained to estimate the Generator's hidden code $c$. In other words, the model estimates the variational lower bound on the mutual information between the latent code $c$ and generated outputs ($\lambda L_I (G, Q)$) with a hyperparameter $\lambda$. In the ciwGAN architecture, the latent code $c$ is a one-hot vector; in the fiwGAN architecture the latent code is a binary code. It has been shown that even if there is a mismatch between the number of lexical items and the number of classes, lexical learning nevertheless occurs in the models \citep{begusCiw}.

During training the Generator learns to generate data such that it increases the Discriminator's error rate and decreases the Q-network's error rate. In other words, the Generator needs to learn to encode unique information into its acoustic inputs, such that the Q-network is able to decode unique information from its generated sounds. The training between the Generator and the Q-network mimics the production-perception loop in speech communication: after training, the Generator learns to generate individual words given a latent code $c$ \citep{begusCiw} and the Q-network learns to classify unobserved words with the same corresponding codes \citep{begus22interspeech}. Since learning is unsupervised, the Generator could in principle encode any information about speech into its latent space, but the requirement to be maximally informative causes it to encode linguistically meaningful properties (both lexical and sublexical information;  \citealt{begus22interspeech}). Such a setting not only replicates the production-perception loop, but is also one of the few architectures featuring traces of communicative intent (between the Generator and the Q-network). Unlike in generative models trained on next sequence prediction or data replication where no communicative intent exists, the training objective between the Generator and the Q-network is to increase mutual information between the latent space and the data such that the Q-network can retrieve the information (latent code) encoded into the speech signal by the Generator.

CiwGAN and fiwGAN have been shown to be highly innovative in linguistically interpretable ways \citep{begusCiw,begus2020identity}. For example, the Generator produces new words or new sound sequences that it never accesses during training. Crucially, the Generator never directly accesses the data: it learns by generating data from noise such that the Discriminator fails to distinguish real and generated data. In this respect, it mimics learning by imitation in human language (rather than replication as is the case with variational autoencoders; \citealt{kingma14}).

Three models are trained for the purpose of this paper, and two models are pretrained from \citet{begusCiw}. The models were trained for approximately 16 hours on a single GPU (NVIDIA 1080Ti). For guidelines on the choice of the number of steps during training, see \citet{begusCiw}. We use the TIMIT \citep{timit} database for training. The number of parameters is given in \ref{sec:appendix}. We take the standard hyperparameters (from \citealt{donahue19} and \citealt{begusCiw}). Because the outputs are salient, we combine transcriptions performed by the authors with transcriptions performed by Whisper \citep{whisper}.

\subsection{Experimental Design}
 The training dataset consists of sliced lexical items from the TIMIT database of spoken English \citep{timit}, such that each item is a single spoken word.

 In order to test robustness of spontaneous concatenation across different models and settings, we conduct five experiments with varying parameter choices. We alter the number of training steps, training words,  word-types, and tokens, randomness of padding, length of output and model architecture. The experiments are summarized in Table \ref{tab:parameters}. 

Models 1 and 2, the \textit{one-second one-word} models, are pre-trained models from \citet{begusCiw}, which we test for signs of concatenation. As specified in \citet{begusCiw}, the Generator and Discriminator are trained on 1.024s segments of audio. Data is never left-padded (only right padded), which controls for the effect of padding on concatenation. Additionally, all training examples contain exactly one word. It is possible that the Generator would learn that silences can occur on both sides of the lexical item in training, and then inverse silences with lexical items which would result in concatenation. The right-padding experiments control for this possibility.

Model 1 uses 5 lexical items with ~600 tokens each: \textit{oily, rag, suit, year} and \textit{water}. Model 2 is trained on 10 lexical items, namely \textit{dark, water, oily, year, greasy, like, carry, ask, rag, suit}. The smaller number of lexical items chosen for training is advantageous from the perspective of modeling language acquisition as it mimics the stage when language-acquiring children have fewer lexical items in their inventory. These particular lexical items were chosen for training because they are well represented non-function words in the TIMIT database \citep{timit}.

Model 3 (\textit{simple two-second one-word}) is another one-word experiment trained on \textit{box, greasy, suit, under}, and \textit{water}. Here, each item is represented by a single token (to test how the model performs with less variability during training) and randomly padded with silence to a length of 2s to produce 100 distinct training instances for each class, for a total of 500 instances used in training (the \textit{two-second one-word} experiment). 

Model 4 (\textit{complex two-second one-word}) is trained with similar hyperparameters as Model 3, but in a more complex setting where each lexical item is represented by approximately 600 distinct tokens. Each token corresponds to a distinct utterance, as found in the TIMIT dataset. With this experiment, we test the ability of the model to learn in a more complex environment. Again, for this experiment each data sample is randomly padded with silence to a length of 2s.

\begin{table}[]
    \centering
    \begin{tabular}{c|ccccccc}\hline\hline
         Model&Words per file& Length  of output & Padding  & Types & Tokens & Steps &Architecture \\\hline
        1&1&1.024s&right&5& $\sim$600&8,011&ciwGAN\\
                        2&1&1.024s&right&10& $\sim$600&7,678&ciwGAN\\
         3&1&2.048s&random&5& 1&8,956&ciwGAN\\
         4&1&2.048s&random&5& $\sim$600&18,842&ciwGAN\\
        5&2&2.048s&random&3& 1&9,166&fiwGAN\\

        \hline\hline
    \end{tabular}
    \caption{An overview of parameters used in the five models. }
    \label{tab:parameters}
\end{table}

    Model 5 (\textit{two-second two-word}), is trained with 3 lexical items with a single token each: \textit{greasy, suit}, and \textit{water}.  Each training example is 2.048s in length and generated by random padding. We generate training examples that contain each combination of two lexical items (i.e.~\textit{greasy, suit}, and \textit{water} alone, \textit{greasy} followed by \textit{water}, \textit{water} followed by \textit{greasy} etc.). We withhold the combination \textit{suit/greasy}, such that they do not appear together in the training set in any order. We generate 100 examples for each combination of lexical items, as well as 100 single-word examples for each lexical item, to produce a final training set of 700 examples. 

For the one-word models, we use the ciwGAN architecture with one-hot encoding and the number of levels equal to the number of types, such that each distinct wordtype in the training corpus can be represented with a unique one-hot vector. In the two-word model, we use a modified fiwGAN (binary code). The binary code is limited to three bits, but each code can have up to two values of 1 (e.g.~[1,0,0] and [1,1,0]).

Generated audio files and models' checkpoints are available at the open source repository: \url{https://doi.org/10.17605/OSF.IO/PRZUQ}.

\section{Results}

We use the technique proposed in \citet{begus19} to analyze the relationship between linguistically meaningful units and latent space. According to this technique, setting individual latent space variables to values outside of the training range reveals the underlying linguistic value of each variable. We address the following research questions based on the five experiments previously described: (i) Are negative latent-code values correlated with concatenated outputs? (ii) Is concatenation an artifact of the training process or a model-generalizable phenomenon? (iii) Can a model trained on both one-word and two-word examples generalize to withheld combinations of lexical items? (iv) Does the model exhibit compositionality when it generates concatenated outputs?

\subsection{Lexical Learning}
In the ciwGAN architecture, the Generator learns to associate each unique one-hot code with a unique lexical item in a fully unsupervised and unlabeled manner \citep{begusCiw}. The Generator's input during training is a one-hot vector with values 0 or 1. For example, the network learns to represent \textit{suit} with $[1,0,0,0,0]$. To test this observation, we set the one-hot vector to values outside the training range (e.g.~$[5, 0, 0, 0, 0]$), which overrides lower-level interactions in the latent space \citep{begus19}. Through this procedure, we can verify that lexical learning takes place in Model 1. In Model 1, the latent code $[5, 0, 0, 0, 0]$ causes the Generator to output \textit{suit} at near categorical levels (9 times out of 10), revealing the underlying value of the code. As further examples, we find that $[0,5,0,0,0]$ encodes \textit{year} (8 times out of 10) and $[0,0,5,0,0]$ encodes \textit{water} (10 times out of 10).

However, the models with random padding, namely Models 3, 4, and 5, display worse lexical learning. The disadvantage of the randomly padded models is that the one-hot code $c$ could also encode the position of the words in the output rather than wordtypes. The advantage of testing concatenation on right-padded models only is that the model receives no evidence for lexical items in the second part of the input, which could in principle promote concatenation.   Therefore, for the purpose of testing compositionality with lexical learning and concatenation, we focus on the right-padded model only, where robust lexical learning in the one-hot codes $c$ has been documented \citep{begusCiw}.
  
\subsection{Concatenated outputs can be reliably elicited with negative-value latent codes
}

\label{sec:oneword}

\begin{figure}
    \centering
    \includegraphics[width=.4\textwidth]{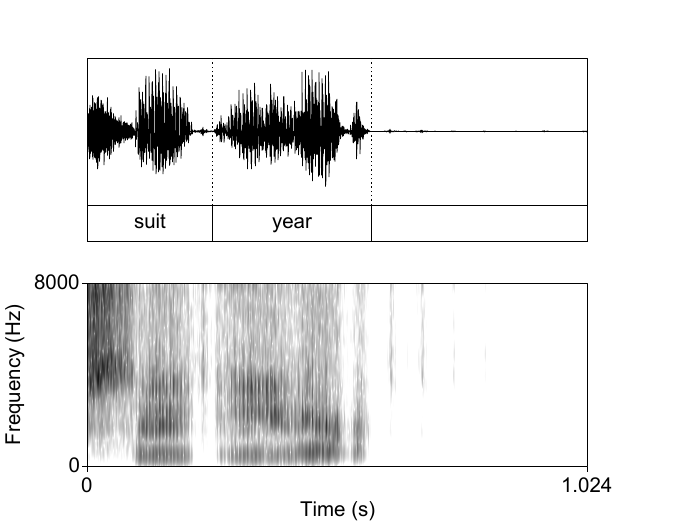}
     \includegraphics[width=.4\textwidth]{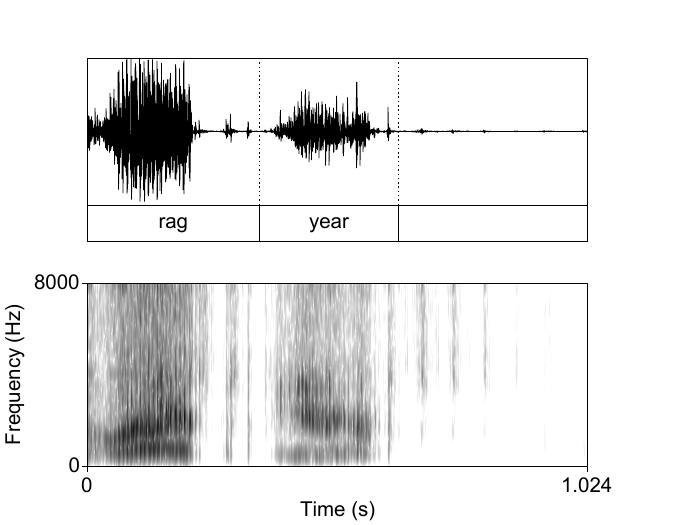}
    \caption{The \textit{suit year} (\textbf{left}) output and the  \textit{rag year} (\textbf{right}) from the one-second one-word model. All spectrograms are created in Praat \citep{boersma15}.}
    \label{fig:suityear}
\end{figure}

We observe a robust pattern in Models 1 and 2, the \textit{pretrained one-word one-second} models: the model Generators produce two-word outputs when the one-hot values are set to negative values outside of the training range. This occurs despite the fact that these models only receive training examples with a single word each. For example, when the latent code is set to $[0,-2,-2,-2,0]$, the Model 1 Generator consistently outputs a two-word output \textit{suit year} (8 times out of 10). For $[-3, -2,-3,-2,2]$, Model 1 consistently outputs \textit{rag year} (8 times out of 10; Figure~\ref{fig:suityear}). These concatenations occur despite the fact that the training data is always left-aligned, the Generator never accesses the data directly and the Discriminator only sees single words.

We also analyze Model 3 (\textit{simple two-second one-word}) for signs of concatenation. While high positive values occasionally yield two-word outputs in this model, negative values are consistently associated with two-word outputs.
For example, $[-50,-50, 0,-50,0]$ (with extreme values) consistently encodes \textit{box greasy} (9 times out of 10), and $[-50,-50, -50,0,0]$ consistently encodes \textit{greasy under} (10 times out of 10).  Positive values of the same codes with extreme values (50) produce completely unintelligible outputs (noise). 

The same phenomenon is observed in Model 4 (in Table \ref{tab:parameters}), which is trained on approximately 600 tokens for each lexical item (for a quantitative analysis of this experiment, see Section \ref{sec:stats} and Figure \ref{fig:numwordsbits}).

\begin{figure}
    \centering
    \includegraphics[width=.43\textwidth]{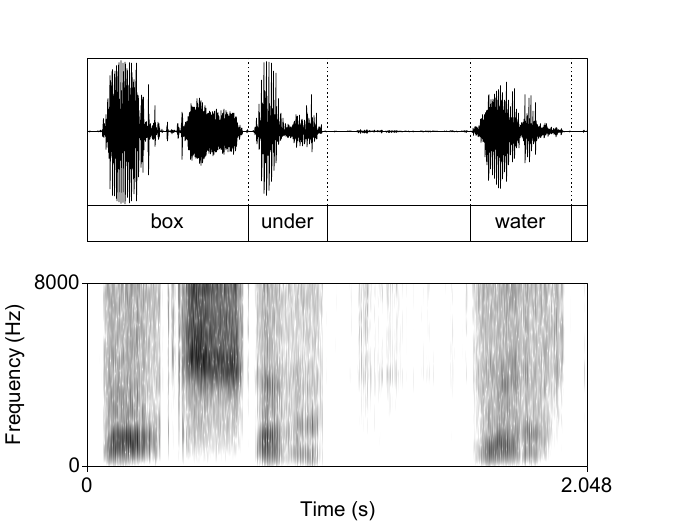}
    \caption{The three-word concatenated output \textit{box under water}. Independently, the second word (\textit{under}) is somewhat difficult to analyze,  but given only five training words, it is clearly the closest output to \textit{under}.}
    \label{fig:boxunderwater}
\end{figure}

 In addition to several two-word concatenated outputs, Model 4 even occasionally generates a three-word concatenated output \textit{box under water} (Figure \ref{fig:boxunderwater}) for the latent code $c$ with all negative values $[-3,-1,-1,-1,-1]$ (2 times out of 10).

\subsubsection{Statistical Analysis}
\label{sec:stats}

\begin{figure}
    \centering
    \includegraphics[width=0.98\textwidth]{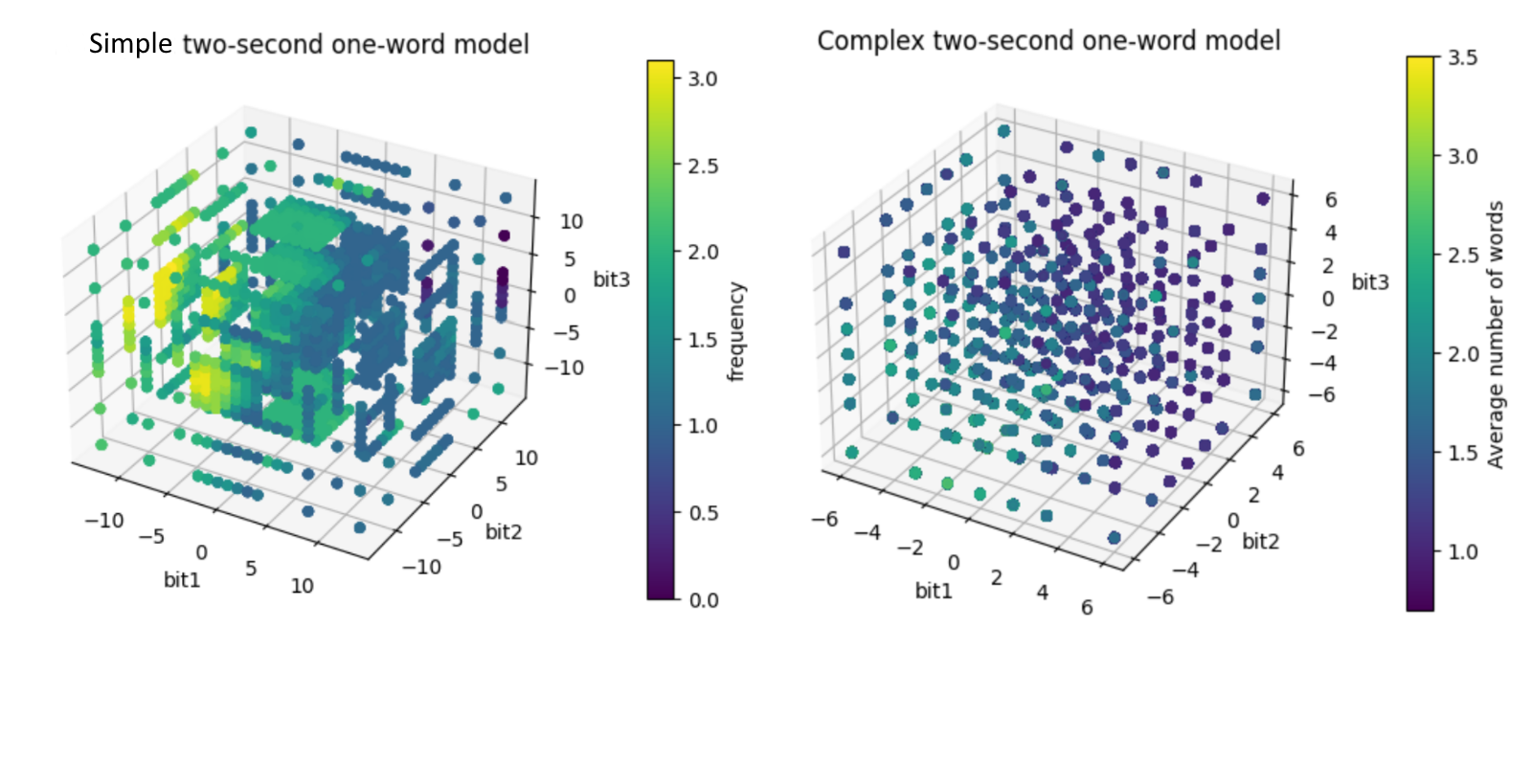}
    \caption{The average number of words generated by the two one-word two-second models are plotted as a function of the first 3 bits of the latent code. The remaining two bits of the latent code are maintained at a value of 0. As in the other trials, each bitstring was tested with 10 sets of latent space values.}
    \label{fig:numwordsbits}
\end{figure}

We test the observation that negative-value latent codes produce concatenated outputs more rigorously.  We systematically generate samples where each of the 5 bits is interpolated from the set of values $\{-6, -3, 0, 3, 6\},$ and all permutations of bits are tested ($5^5$). Each permutation is tested with 10 sets of latent space values (constant across different permutations), for a total of $31,250$ samples. We automatically annotate our outputs using a pretrained Whisper-small model \citep{whisper}, which we fine-tune using a dataset consisting of 400 model outputs manually annotated by the authors. Whisper is used to annotate for single (failure; 0) vs.~multiple (success; 1) word outputs. We estimate Whisper's accuracy on labeling single- and multiple-word outputs at 88\% (based on author's annotations; Table \ref{whisper-accuracy}).

\begin{figure}
    \centering
\includegraphics[width=1\textwidth]{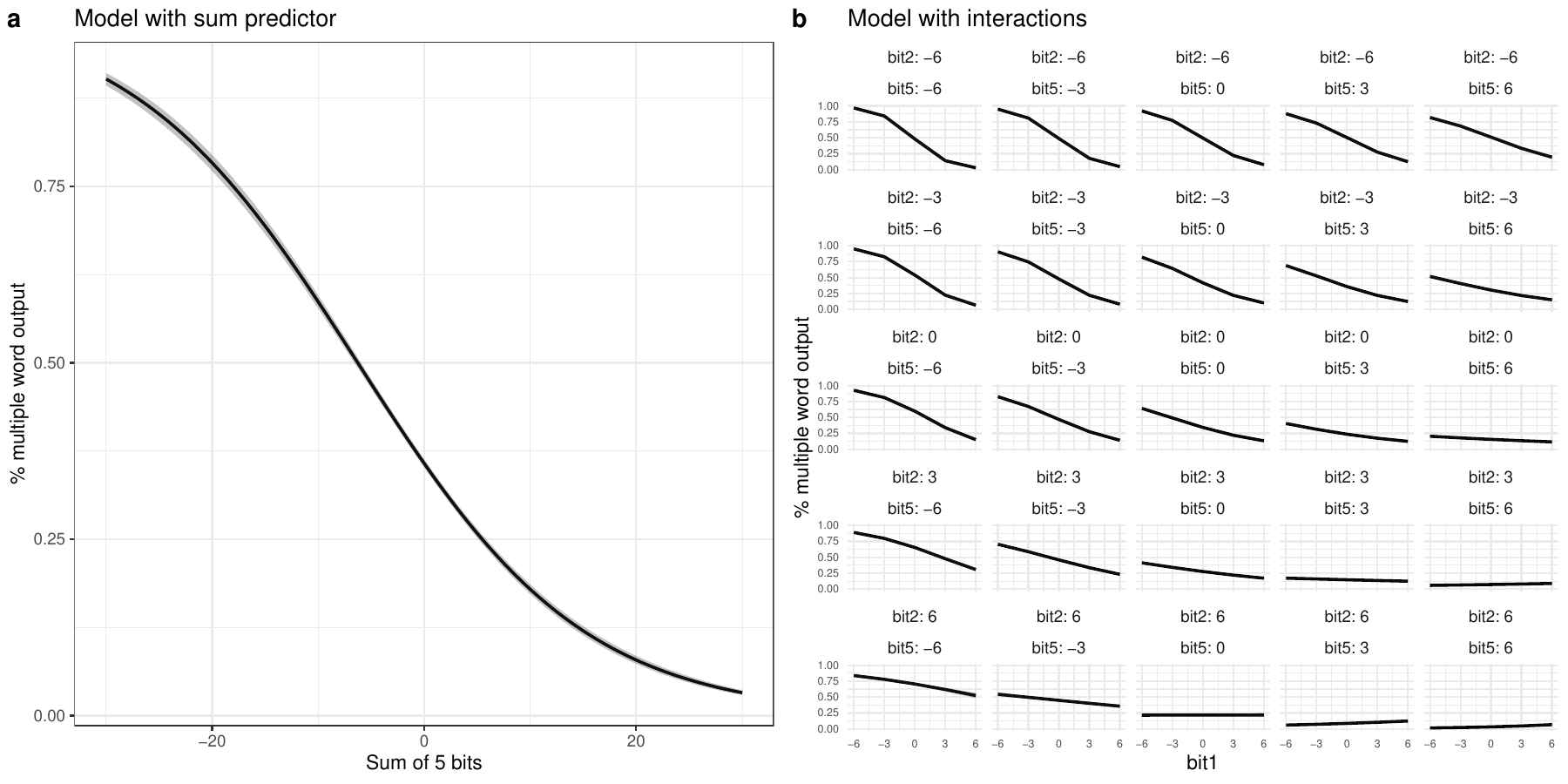}

    \caption{(\textbf{a}) Predicted values of the logistic regression mixed effects model with the proportion of two-word outputs (one-word output = failure, two-word output or more = a success) as the dependent variable and sum of bits 1--5 as the predictor with 95\% confidence limits (Model 3 in Table \ref{tab:parameters}). Outputs with no transcribed words were removed.  The random effect structure involves  random intercepts for each of the 10 unique random latent space samples $z$ as well as random slope for sum of bits. The values were obtained with the \textit{effects} package \citep{effects1,effects2}. 
(\textbf{b}) Predicted values for one (\texttt{bit1 * bit2 * bit5}) of the many interactions of the logistic regression mixed effects model with the proportion of two-word outputs (one-word output = failure, two-word output = a success) as the dependent variable and individual bits with all interactions (including the five-way interaction) as the predictors with 95\% confidence limits (the model is set up as \texttt{two or more words $\sim$ bit1 * bit2 * bit3 * bit4 * bit5 + (1|z)}). The random effect structure involves only the random intercept for each of the 10 unique random latent space samples $z$. The values were obtained with the \textit{effects} package \citep{effects1,effects2}. Model estimates are given in Table \ref{estimates}. Not all interactions show the same effect.
}
    \label{fig:var5single2srun1model1ggplot}
\end{figure}

Raw data in Figure \ref{fig:numwordsbits} shows that Model 3 generates between 0 and 3 words for each $z$ vector, generating substantially more words at negative bit values. The data was fit to a logistic regression mixed effects model (details of the model in Figure \ref{fig:var5single2srun1model1ggplot}). There is a clear and significant negative relationship between the sum value of bits and the probability of a multiple-word output ($\beta=-0.09,   z= -54.42, p  <0.0001$; Figure \ref{fig:var5single2srun1model1ggplot}). The mixed-effect model with interactions between individual bits also shows that negative values correspond to significantly higher proportion of  two-word outputs (Figure \ref{fig:var5single2srun1model1ggplot}, Table \ref{estimates}).

We perform a similar analysis on Model 4. We systematically generate samples where each of the 5 bits is interpolated from the set of values $\{-4, -2, -1, 0, 1, 2, 4\}$, and all permutations of bits are tested $(7^5),$ Again, each permutation is tested with 10 sets of latent space values (constant across different permutations), for a total of $168,070$ samples. All samples were subsequently transcribed using the same fine-tuned Whisper model used in the previous experiments, and the results suggest a similar correlation between negative code values and the proportion of two-word outputs (Figure 4).

The data of this experiment was fit to a logistic regression model which again shows a significant relationship between the proportion of multiple-word outputs (the dependent variable) and the sum of all the latent code bits (the independent variable) ($\beta=-0.14,   z= -167.5 , p  <0.0001$). All details of the model and plotted estimates are in Figure \ref{fig:var5single2srun600ggplot}.

\subsection{Concatenation across Model Configurations
}

Of particular interest is the robustness of our observation that negative-value latent codes produce concatenated outputs. We show that this pattern persists across multiple training configurations, suggesting that concatenation is an innate property of the model architecture as opposed to an artifact of particular hyperparameter settings. 

First, we observe that the one-word models (Models 1-4) are trained exclusively with one-word examples. Only the Discriminator sees these training examples; the Generator, which is responsible for concatenation, is trained only from Discriminator feedback. Similarly, in Model 5, the Generator learns to produce the \textit{suit/greasy} combination that is withheld from training data. Thus, we argue that concatenation is an emergent property of the Generator, as opposed to something learnable from training data.

We also consider the alignment of words within each audio segment as a potential cause of concatenation. We find that Models 1 and 2, which are trained on exclusively left-aligned data, exhibit concatenation. Generating concatenated outputs in this setting is more challenging, as the Generator must learn to right-pad the second word in the concatenated sequence, as opposed to directly superimposing two productions already within the Generator's vocabulary. On the other hand, Models 3 and 4, which are randomly padded, also exhibit concatenation; therefore, we conclude that concatenation as a property does not depend on the padding of training data.

\begin{figure}
    \centering
    \includegraphics[width=0.4\linewidth]{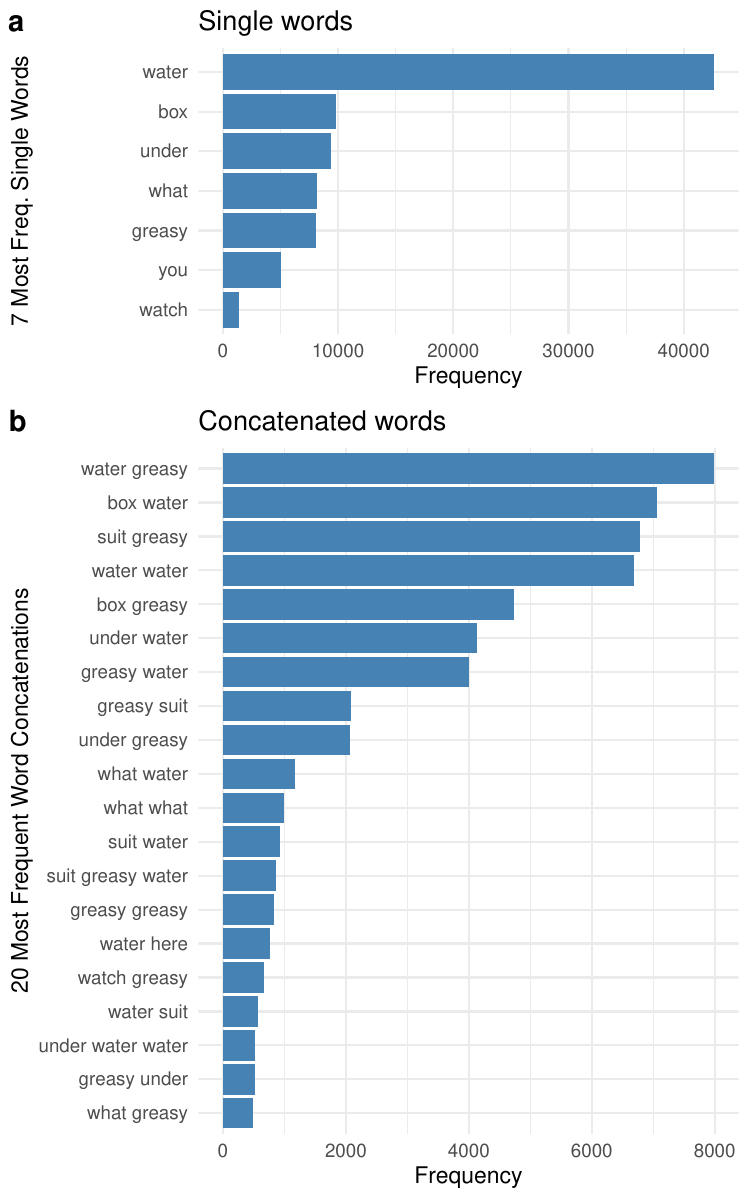}
    \caption{Counts of word types as transcribed by fine-tuned Whisper. (a) Counts of 7 most frequent outputs that were transcribed as including a single word by the fine-tuned Whisper.  (b) Counts of 20 most frequent outputs that were transcribed as including two or more words by the fine-tuned Whisper.  }
    \label{fig:top7}
\end{figure}

We find that concatenated words are well distributed across word types, instead of being concentrated on a few specific examples. We consider the same dataset of $168,070$ transcribed outputs generated by Model 4 (with all permutations of code values $\{-4, -2, -1, 0, 1, 2, 4\}$) as the analysis performed in Section \ref{sec:stats}. The fine-tuned Whisper model used to perform the transcriptions has approximately 54\% word-level accuracy compared to human evaluations, but this criterion is conservative as only perfect matches are counted as success, and the majority of errors occur when Whisper transcribes to words outside of the training set. In single word outputs, the word \textit{water} is overrepresented with the rest of the lexical inventory relatively equally represented. In concatenated word outputs, the distribution of word types is also relatively uniform. The five training words are well represented in both the first and second positions, which suggests that word concatenation is not an idiosyncratic property of a few word types or the result of an acoustic artifact. Figure \ref{fig:top7} illustrates the most common types for single-words and for concatenated outputs.

\subsubsection{Single Batch Model}

We also consider the possibility that spontaneous concatenation arises solely due to the models being trained in a batched training configuration, where different samples in the same batch could be concatenated when their gradients are combined into a single training update. To demonstrate that spontaneous concatenation occurs even in the absence of batching, we train an additional model using the same dataset and hyperparameters as Model 4, but with the batch size set to $1.$ We trained this model for 41,922 steps. It is well-known that batch size can affect training outcomes \citep{lin22}. Because of the small batch size, the words generated by this model are qualitatively poorer, which prevents an automated Whisper analysis of the outputs. However, we continue to consistently observe multiple-word outputs in the negative latent space, indicating that this behavior is not solely attributable to batching. We randomly generated 10 examples from this model with the latent code set at [-1, -1, -1, -1, -1]. All 10 examples feature two or more word sequences, although the audio quality is too low to identify the words. Waveforms and spectrograms of the ten examples are given in Figure \ref{fig:oneBatch}.

\subsection{Generalization to Withheld Combinations
}

Model 5 is trained with both one-word and two-word examples. The model is only trained on three lexical items and their pairwise combinations, except for the withheld \textit{suit/greasy} combination. We find that the Generator from Model 5 consistently outputs the unobserved \textit{greasy suit} for $[15,0,0]$ (17 times out of 20), which suggests the network learned this unobserved combination as one of the possible sequences and encoded it with a one-hot value. 

\begin{figure}
    \centering
    \includegraphics[width=.5\textwidth]{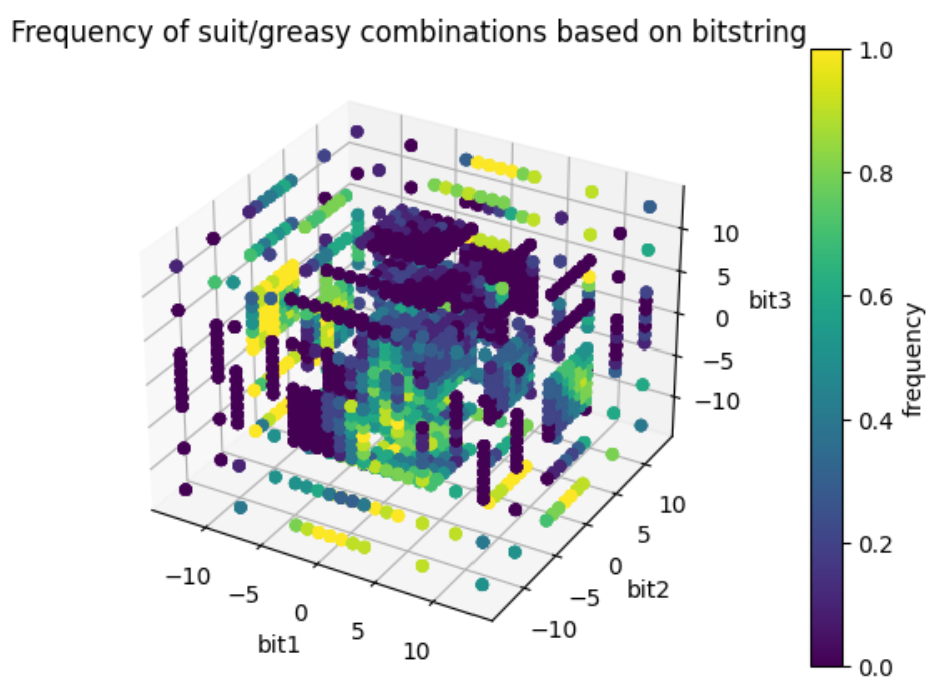}
    \caption{The frequency of observed \textit{suit}/\textit{greasy} pairs is plotted as a function of the latent code, for the fiwGAN two-word model. Samples were generated by varying each bit between the values [-13, -8, -3, -2, -1, 0, 1, 2, 3, 8, 13], taking all possible permutations ($11^3$). Each bitstring was tested with 10 sets of latent space values, and annotations were performed automatically using Whisper.}
    \label{fig:suitgreasyplot}
\end{figure}

To evaluate the presence of the withheld \textit{greasy suit} pair systematically, we conduct an exploration of the latent space (all permutations where each variable takes on values in  $\{$-13, -8, -3, -2, -1, 0, 1, 2, 3, 8, 13$\}$) in Figure \ref{fig:suitgreasyplot}. As before, we evaluate 10 different latent space values for each code, and transcribe outputs automatically with Whisper. The withheld \textit{suit/greasy} combinations occur most often for codes around [-10, 0, 0], or otherwise when at least one bit is negative (Figure \ref{fig:suitgreasyplot}), suggesting that the model has associated the unobserved \textit{suit/greasy} combination with that code. While there are localized pockets of the concatenated outputs in the positive area of the latent space, the concatenated outputs are reliably present in the negative areas. 
It appears that the negative values of the latent code $c$ again encode unobserved novel combinations. The representations, however, are not yet compositional: as will be shown in Section \ref{disinhibition}, compositionality emerges when a negative value starts representing a single but concatenated word (see Equations \ref{eq3} and \ref{step2}).

\subsubsection{Repetition}
\begin{figure}
    \centering
    \includegraphics[width=.4\textwidth]{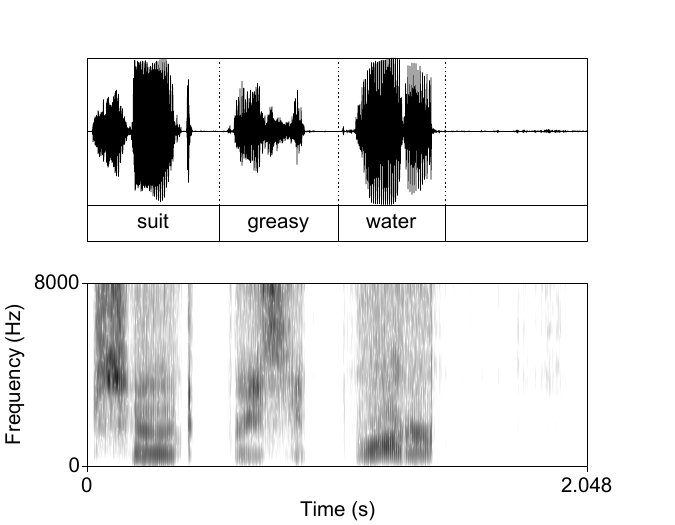}
\includegraphics[width=.4\textwidth]{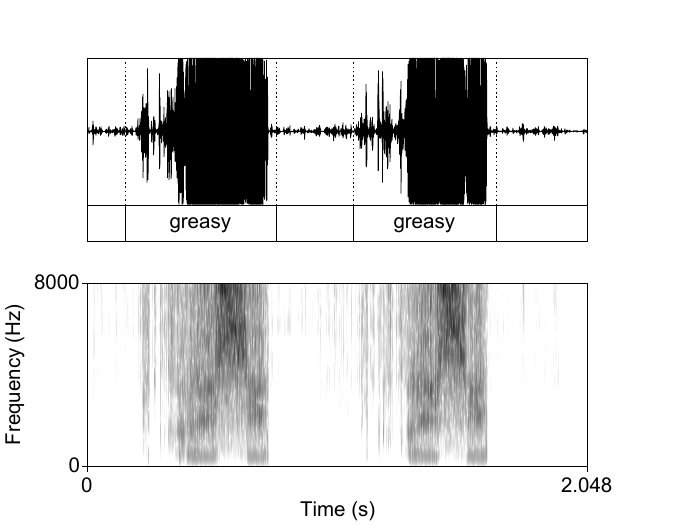}

    \caption{The \textit{suit greasy water}  output from the two-second two-word model (\textbf{top}). The \textit{greasy greasy}  output from the two-second one-word model (\textbf{bottom}). }
    \label{fig:suitgreasywater}
\end{figure}

In addition to two-word concatenation and embedding of words into novel combinations, we also observe outputs with repeated words in all our trained models. The training data never includes the same word repeated, yet the models frequently include repeated words. For example, the two-second one-word model consistently outputs \textit{greasy greasy} for $[0, 0, -40, 0,0]$ (7 times out of 10; Fig.~\ref{fig:suitgreasywater}). Of the 31,250 samples annotated in the analysis, 2753 (8.8\%) contained repeated words, suggesting that the model systematically learned this behavior. This is significant because repetition or reduplication is one of the most common processes in human language and language acquisition \citep{berent16,dolatian20}.

\subsection{Semi-Compositional Concatenation
}
\label{compositionality}


Compositionality has long been a focus of computational models (\citealt{kirby01,andreas18,chaabouni20,rita22}; for a review, see \citealt{rita24}). Most evaluated models, however, are trained on already compositional data or operate with symbolic units rather than with raw data. Our model is trained on non-compositional single unit words. Concatenation and compositionality need to self-emerge in our setting.

Experiments in Section \ref{sec:oneword} suggest that negative values in the latent space result in spontaneously concatenated outputs. Here, we show that these spontaneously concatenated outputs show traces of compositionality. 

CiwGAN has been shown to learn lexical items \citep{begusCiw} in a fully unsupervised manner, associating each one-hot code with a distinct lexical item. For example, when Model 1 is trained on approximately 600 tokens of five lexical items, it learns to associate [1, 0, 0, 0, 0] with the word \textit{suit},  [0, 1, 0, 0, 0] with \textit{year}, [0, 0, 1, 0, 0] with \textit{water},  [0, 0, 0, 1, 0] with \textit{oily}, and  [0, 0, 0, 0, 1] with \textit{rag} \citep{begusCiw}. Lexical learning thus emerges in a fully unsupervised way, only from the requirement that the networks can recover the latent code from synthesized audio. 

Single lexical items represented with unique one-hot codes can be causally forced in the output at near categorical levels, by setting the one-hot code to extreme values outside the training range of [0,1]. For example, setting the code to [0, 0, 0, 0, 2], the network outputs \textit{rag} in 98\% of outputs. 

\begin{figure}
    \centering
    \includegraphics[width=0.8\linewidth]{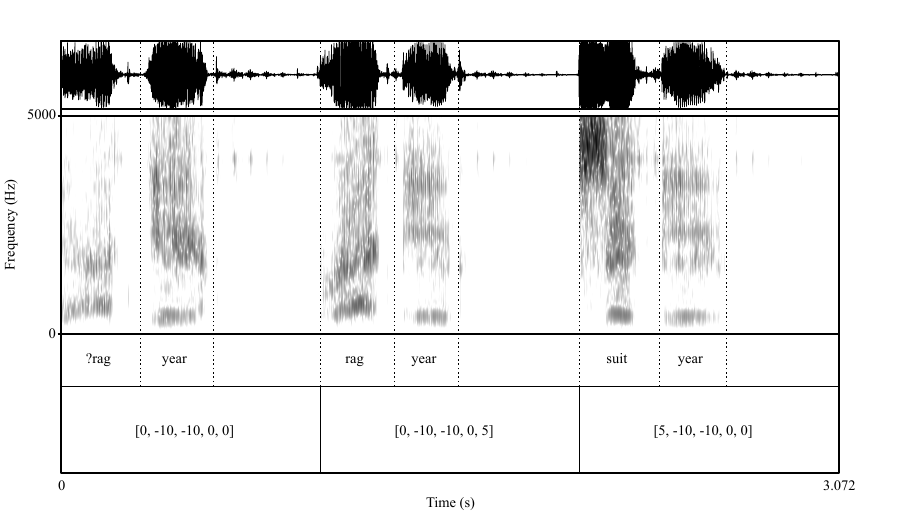}
    \caption{Waveform and spectrograms (0--5kHz) of three generated outputs with latent codes [0, -10, -10, 0, 0], [0, -10, -10, 0, 5], and [5, -10, -10, 0, 0].}
    \label{causalcompositionality}
\end{figure}

Exploration of the negative extreme values in the one-hot code $c$ shows that some values represent a specific single word or a word concatenated with another, non-specific word that is difficult to acoustically identify. For example, Model 1 consistently outputs the word \textit{year} together with another non-specific word when its latent code is set to [0, -10, -10, 0, 0]. Figure \ref{causalcompositionality} illustrates one such example. We generate five other outputs (audio samples available at the data repository link), but they are nearly identical to the one displayed in Figure \ref{causalcompositionality}. 

The first word in this output (Figure \ref{causalcompositionality}) can be transcribed as \textit{rag}, but has low quality. The second word can be transcribed as \textit{year}. Figure \ref{causalcompositionality}) shows that [0, -10, -10, 0, 0] represents the word \textit{year} concatenated with another word with a weaker audio profile.

This becomes more apparent when we combine [0, -10, -10, 0, 0]  with the code that represents the word \textit{suit} ([1, 0, 0, 0, 0]) and set both to extreme values ([5, -10, -10, 0, 0]). When the latent code $c$ is compositionally combined in such a way, we get a clear case of a compositional concatenation in the output: \textit{suit year}. We generate 5 outputs with different latent space values, and find that they all contain the \textit{suit year} concatenation. When we combine [0, -10, -10, 0, 0] with the value that represents \textit{rag}, we get a clear output containing \textit{rag year}. We can thus causally intervene and compositionally force the desired lexical concatenated items in the outputs. This happens in Model 1, which was only trained on single right-padded word inputs. While such a clear causal relationship between the compositionality of the latent code and the compositionality of the concatenated outputs exists for a subset of codes and not all codes, this is a step towards a compositional representation of concatenation.

Compositionality is even more pronounced in Model 2. The model, trained on ten words, learns to represent words with unique one-hot codes. For example, [1, 0, 0, 0, 0, 0, 0, 0, 0, 0] represents the word \textit{dark}, etc. Again, setting the values of the code to extreme values (in line with \citealt{begus19,begusCiw}) reveals the underlying value of each code and results in a single lexical item in the output at near categorical levels.

We replicate the previous finding from Model 1: setting the latent code to extreme negative values in a subset of cases results in a specific word either surfacing alone or concatenated with another word with a low acoustic profile. For example, in about half of the outputs, the code [0, 0, -1.5, 0, 0, 0, 0, 0, 0, 0] results in a single word \textit{greasy} with a relatively low, but identifiable quality (most of the remaining outputs are not identifiable because of poor acoustics).  When this code is compositionally combined with extreme positive values of codes that represent other words, we get predictable concatenated words of \textit{greasy} together with the words that a specific code represents.  This strongly suggests [0, 0, -1.5, 0, 0, 0, 0, 0, 0, 0] represents \textit{greasy} and that compositionally combining this code with codes that represent other words leads to a predictable concatenation.

\begin{figure}
    \centering
    \includegraphics[width=0.85\linewidth]{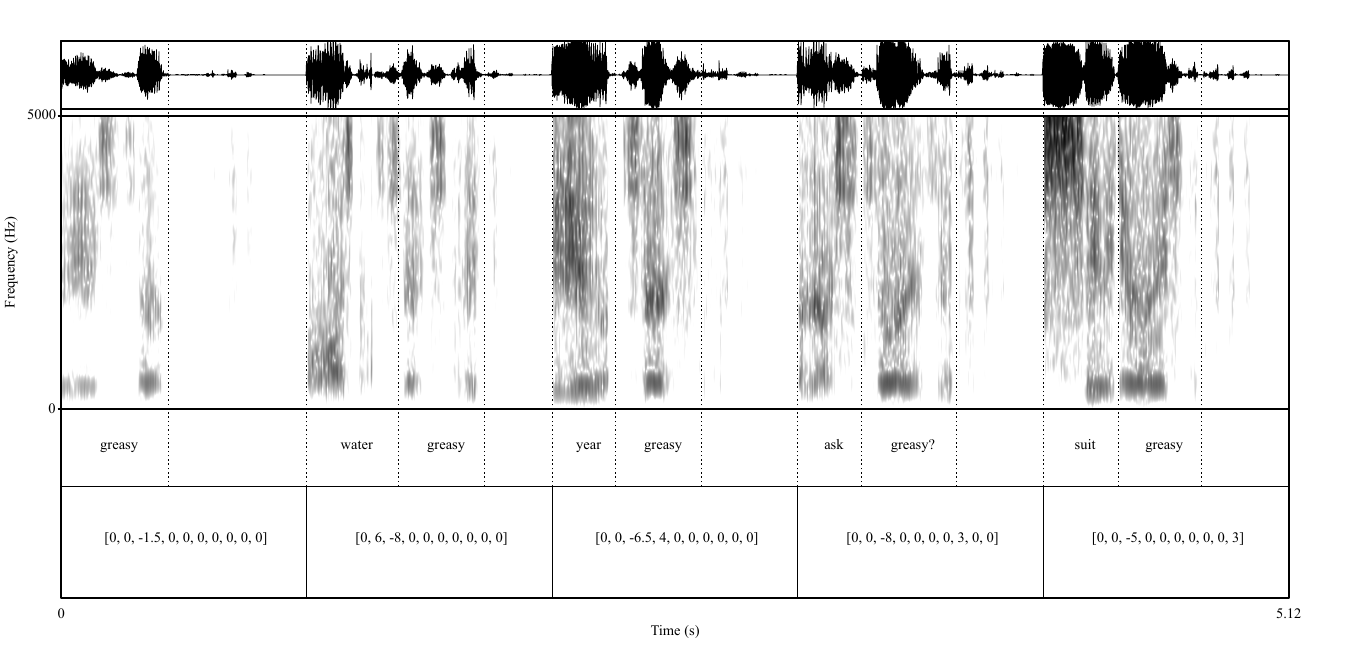}
    \caption{Waveforms and spectrograms (0-5 kHz) of five generated outputs with latent code variables as described in Table \ref{tab:codescomp}. The first output (from left) represents \textit{greasy} with negative values of the latent code $c$. The following outputs have compositional values of the latent code $c$: the negative value associated with \textit{greasy} is combined with the positive value that represent individual lexical items. The outputs contain compositionally concatenated words with predictable word-pairs based on the latent code.}
    \label{fig:watergreasychain}
\end{figure}

For example, code [0, 0, 0, 1, 0, 0, 0, 0, 0, 0] represents \textit{year}. A combination of this word plus the negative code representing \textit{greasy} gives a clearly identifiable  \textit{year greasy} ([0, 0, -6.5, 4, 0, 0, 0, 0, 0, 0]) in 10/10 generated outputs. A combination with the codes representing \textit{ask}, \textit{suit}, and \textit{water} results in \textit{ask greasy}, \textit{suit greasy}, and \textit{water greasy} in 10/10 or 9/10 outputs.  Table \ref{tab:codescomp} illustrates the full compositionality of the concatenated outputs and Figure \ref{fig:watergreasychain} illustrates the causally compositional outputs with waveforms and spectrograms.

\begin{table}[]
    \centering
\scalebox{.7}{    \begin{tabular}{cc|cc|rr}
\hline\hline
       Positive code  & Word & Negative Code & Word & Positive \& Negative Combined & Concatenated Words  \\\hline
       [0, 0, 0, \colorbox{green!30}{1}, 0, 0, 0, 0, 0, 0]  & year & [0, 0, \colorbox{red!30}{-1.5}, 0, 0, 0, 0, 0, 0, 0] & greasy & [0, 0, \colorbox{red!30}{-6.5}, \colorbox{green!30}{4}, 0, 0, 0, 0, 0, 0]& year greasy\\

           [0, 0, 0, 0, 0, 0, 0, 0, 0, \colorbox{green!30}{1}]  & suit & [0, 0, \colorbox{red!30}{-1.5}, 0, 0, 0, 0, 0, 0, 0] & greasy & [0, 0, \colorbox{red!30}{-5.0}, 0, 0, 0, 0, 0, 0, \colorbox{green!30}{3}]& suit greasy\\
           
           [0, 0, 0, 0, 0, 0, 0, \colorbox{green!30}{1}, 0, 0]  & ask & [0, 0, \colorbox{red!30}{-1.5}, 0, 0, 0, 0, 0, 0, 0] & greasy & [0, 0, \colorbox{red!30}{-8.0}, 0, 0, 0, 0, \colorbox{green!30}{3}, 0, 0]& ask greasy\\

              [0, \colorbox{green!30}{1}, 0, 0, 0, 0, 0, 0, 0, 0]  & water & [0, 0, \colorbox{red!30}{-1.5}, 0, 0, 0, 0, 0, 0, 0] & greasy & [0, \colorbox{green!30}{1}, \colorbox{red!30}{-8.0}, 0, 0, 0, 0, 0, 0, 0]& water greasy\\

                         [0, 0, 0, 0, 0, 0, 0, 0, 0, \colorbox{green!30}{1}]  & suit & [0, 0, 0, 0, 0, 0, \colorbox{red!30}{-8.0}, 0, 0, 0] & X carry & [0, 0, 0, 0, 0, 0, \colorbox{red!30}{-12}, 0, 0, \colorbox{green!30}{7}]& suit carry\\
                         
    [0, 0, 0, 0, 0, 0, 0, \colorbox{green!30}{1}, 0, 0]  & ask & [0, 0, 0, 0, 0, 0, \colorbox{red!30}{-8.0}, 0, 0, 0] & X carry & [0, 0, 0, 0, 0, 0, \colorbox{red!30}{-24},  \colorbox{green!30}{5}, 0 ,0]& ask carry\\                   
\hline\hline
    \end{tabular}}
    \caption{Compositionality of codes and their outputs in Model 2.}
    \label{tab:codescomp}
\end{table}

Compositional outputs, while not available for all lexical items, are not limited to \textit{greasy} only. For example, code [0, 0, 0, 0, 0, 0, -8, 0, 0] represents \textit{carry} in concatenation with a word that has unrecognizable acoustics (potentially close to \textit{ask}). When this code is causally combined with the code representing \textit{suit}, the output is a concatenated \textit{suit carry}. While the acoustics are not as clear as in the \textit{greasy} examples, the concatenations are clearly compositional. When the code representing concatenated \textit{carry} is compositionally combined with the code representing \textit{ask}, the output results in a clearer \textit{ask carry}. Table \ref{tab:codescomp} illustrates this compositionality. It appears that the lexical item represented with the positive code appears as the first part of the concatenated output, whereas the lexical item represented with the negative code appears as the second part of the output.

All generated outputs with the exception of the one representing \textit{greasy} are nearly identical across 10 generated repetitions and are available in the data repository. Spectrograms in Figures \ref{fig:watergreasychain} and \ref{fig:askCarry} are representative of the outputs' acoustic content.

Thus, we find that both Models 1 and 2, which are independently trained with different datasets and parameters, learn lexical items and learn to concatenate them compositionally at least in a subset of possible combinations. In a model that was trained exclusively on single word inputs, we have uncovered compositionality  for at least six concatenated pairs. For the word \textit{greasy}, four out of ten possible concatenated outputs were found. For \textit{carry}, two out of ten possible were found.

\subsection{Limitations}

This paper models concatenation of acoustic lexical items with traces of compositionality. Syntax is substantially more complex than concatenation and compositionality \citep{berwick19}. 
In our models, meaning is represented abstractly with a one-hot encoding. While one-hot code can abstractly represent multimodal meaning, a more realistic model would have to contain referential meaning. 

We also train the network on a relatively small number of lexical items (5) and a small number of tokens (from one to approximately 600). The small number of lexical items is representative of the earliest stages of language acquisition when the number of lexical items is highly limited \citep{bates94}. Human syntax, however, operates on a substantially larger number of elements. How this approach scales up is left for future work.

\section{Discussion}
\label{discussion}

\subsection{Artificial disinhibition}
\label{disinhibition}

Here we propose a potential neural explanation for why negative values result in outputs with concatenated words. We propose that spontaneous semi-compositional concatenation results from artificial disinhibition. Disinhibition (or excitation that results from inhibition of inhibitory neurons; \citealt{pi13,pfeffer13}) is a well-established biological neural process. Negative values in the latent space can be interpreted as artificial analogs to inhibitory neurons. If inhibitory neurons operate on another set of inhibitory neurons, the result is excitatory (the so-called disinhibition; \citealt{pi13,pfeffer13}).  

\begin{figure}
    \centering
    \includegraphics[width=0.75\linewidth]{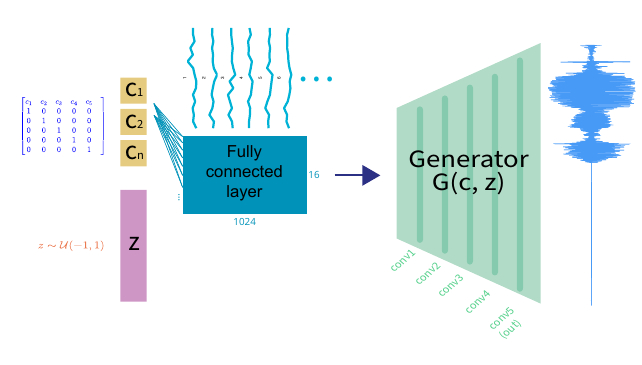}
    \caption{The structure of the Generator. Each variable in the latent space (5 code variables $c$ and 95 uniformly distributed variables $z$) are fully connected to all dimensions of the fully connected layer ($16\times1024$). The figure only illustrates a few connections. The fully connected layer contains 1024 time series data with the dimension of 16. The 16 dimensions correspond to the time domain in the audio output. The output of the fully connected layer gets upsampled to audio output in the final layer via 5 convolutional layers. The six time series data above the Fully connected layer box illustrate the first six out of 1024 time series data. In this paper, we analyze averages of 1024 time series data in the Fully connected layer.}
    \label{fig:fullyconnectedsketch}
\end{figure}

The first layer in our convolutional Generator is the fully connected layer. In this layer, all 100 latent variables (code variables $c$ and random variables $z$) are connected to all variables in the fully connected layer with weights that get updated during training. The fully connected layer is of the shape $1024\times 16$, which means that there are 1024 time series of length 16. The 16 dimensions represent the time domain and get transformed to the time dimension in the final output via the convolutional layers (Figure \ref{fig:fullyconnectedsketch}).

The fully connected layer includes both excitatory artificial neurons (or positive weights that generate words) as well as inhibitory artificial neurons (or negative weights that generate non-activity or silence). The non-activity or silences are likely generated by negative weights which results in negative values of the fully connected layer.  The $c$ variables are always 0 or 1, which means that the Generator is never exposed to negative values during training.  When negative values in the latent space get multiplied with negative weights, however, the resulting output in the fully connected layer will involve  positive values. This inhibition of inhibitory weights (or disinhibition) is the mechanism whereby the Generator can generate multiple words in the output. This process mimics the biological \textit{disinhibition} in human neuroscience.

\begin{figure}
    \centering
    \includegraphics[width=0.8\linewidth]{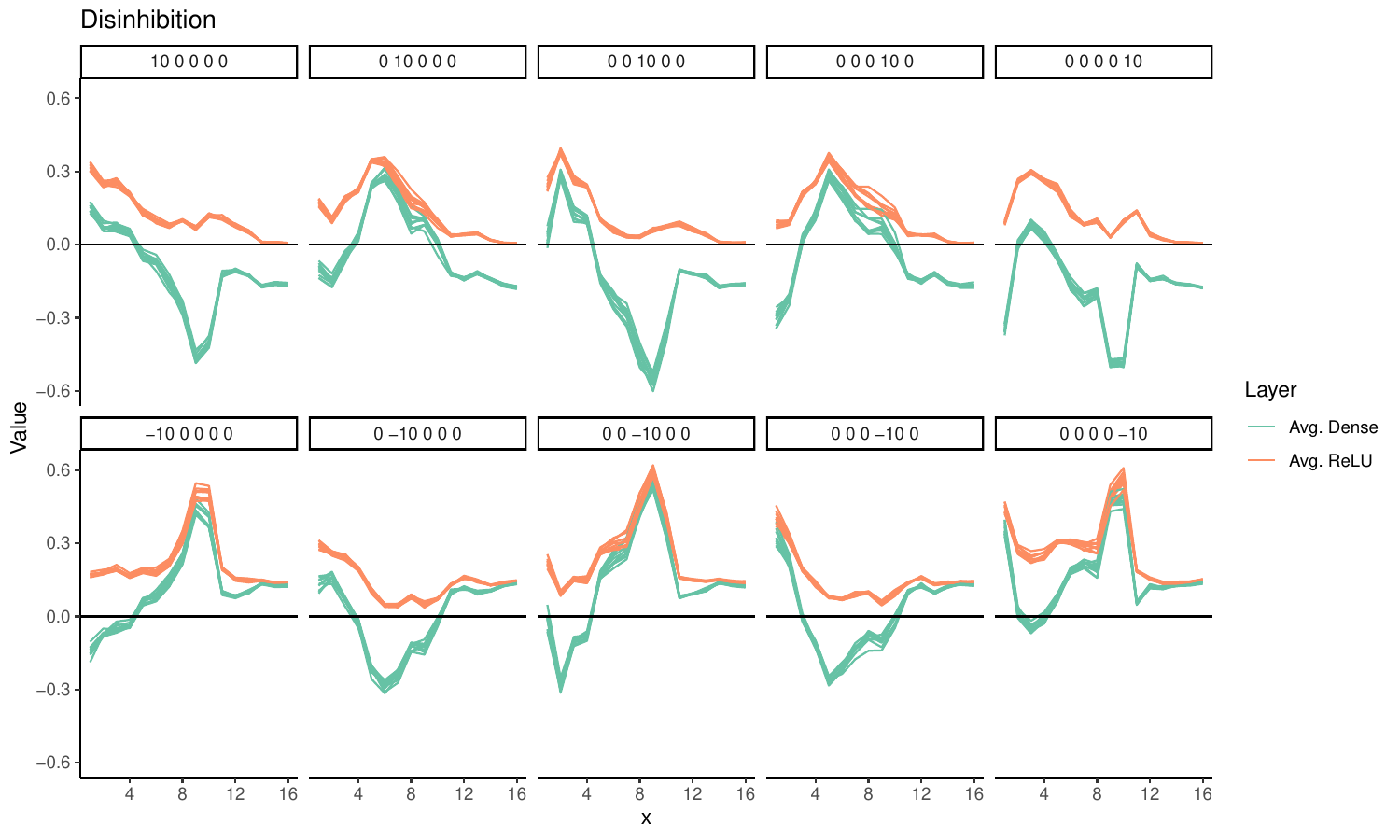}
    \caption{Averaged values of the Dense layer output before (Avg.~Dense) and after ReLU (Avg.~ReLU) for positive and negative extreme values of the latent code $c$.  Each latent code (e.g.~[10, 0, 0, 0, 0] or [-10, 0, 0, 0, 0]) was sampled randomly ten times. }
    \label{fig:disinhibition}
\end{figure}

Figure \ref{fig:disinhibition} illustrates the pathway to how negative values (or disinhibition) result in concatenated outputs. We plot ten outputs from the fully connected layer for each of the five one-hot codes with positive values (with the value of the code set to 10, e.g.~[10, 0, 0, 0, 0]) and the corresponding negative values (e.g.~[-10, 0, 0, 0, 0]) in Model 1. The fully connected (Dense) layer consists of 1024 instances of 16-valued outputs. The 16 values represent the time domain and get transformed to the actual temporal dimension in the generated audio through the convolutional layer. In line with \citet{begusZhou,begusZhou1}, we average each output in the time domain, which yields a single 16-dimension output that summarizes neural activity in the deepest layers in the convolutional network.  We plot both pre-ReLU and post-ReLU values.  

That words are represented by unique shapes in earlier convolutional layers has been shown previously by \cite{begusZhou1}.  Negative values in the latent space cause the fully connected layer to have the inverse of the shape caused by the positive latent values (or disinhibitory values)  as is clear from Figure \ref{fig:disinhibition}. After the ReLU activation (which itself is a biologically plausible operation), the negative or disinhibited values (inverse of the positive values) can cause multiple shapes rather than a single shape with positive values. The shapes in the deeper layers get then transformed into distinct lexical items through a series of convolutional layers \citep{begusZhou,begusZhou1}.

This line of reasoning thus uncovers a possible neural mechanism behind the precursors of syntactic concatenation or even the simplified non-recursive \textit{Merge} operation \citep{chomskyLectures}. In the initial stage, the concatenated output can be represented as a disinhibition, which generates excitatory activity of two or more words as part of a single output. In other words, a concatenated output has a unique representation:  disinhibition that results in activation of two words. This point represents a non-compositional concatenative stage in the evolution/acquisition of syntax. 

As suggested by our experiments in Section \ref{compositionality}, the next development towards compositionality occurs when the disinhibited values start representing a single word concatenated with another word. At this point, a combination of excitation that activates one word and disinhibition that activates another word can result in a causally predictable output that compositionally concatenates the two words.

Disinhibition outlines a potential artificial and biological neural pathway towards concatenated outputs and compositionality in human syntax.

\subsection{Formalizing disinhibition}

\label{formalizingdisinhibition}

Here, we outline how a precursor to the \textit{Merge} operation can empirically and numerically emerge in a fully connectionist model. \citet{chomskyLectures} and \citet{marcolli23}, for example, define Merge ($\mathfrak{M}$) as in \ref{MergeC}, where $\alpha$ and $\beta$ are arguments.

\begin{equation}
    \mathfrak{M}(\alpha,\beta):=
\{\alpha,\beta\}
\label{MergeC}
\end{equation}

Under the minimalist syntactic approach, the set $\{\alpha,\beta\}$ is compositional. In our case, the apparent $\{\alpha,\beta\}$ is represented idiosyncratically as the disinhibitory neuron. The disinhibitory neuron causes the concatenated shapes of two or more words, which are executed by the Generator as two words in the output. We simplify $\alpha$ and $\beta$ to represent lexical items with some associated meaning in the form of one-hot codes. According to the formalism, we can represent the \textit{Disinhibition} ($\mathfrak{D}$) as:

\begin{equation}
    \mathfrak{D}(\gamma)=\{\alpha,\beta\}
    \label{eq3}
\end{equation}

This step does not represent the full extent of compositional concatenation, because the relationship between $\gamma$ and the merged $\{\alpha,\beta\}$ is not decomposable or compositional. However, this step represents the first operation required in the development from the non-concatenative simple signals to complex compound signals of human syntax.  

Based on our experiment in Section \ref{compositionality}, it appears that the step from non-compositionality toward compositionality occurs through an intermediate stage when the negative, disinhibitory values start representing a specific lexical item concatenated with another non-specific word. We have shown that a subset of disinhibited values consistently represent a single lexical item or the specific lexical item concatenated with another non-specific item. Such disinhibitory values are formalized in \ref{step2}.

\begin{equation}
    \mathfrak{D}(\gamma)=\{\alpha,\_\}
    \label{step2}
\end{equation}

Compositional concatenation can now be achieved by joining the values of $\mathfrak{D}(\gamma)$ and $\beta$. 

\begin{equation}
    \mathfrak{C}(\mathfrak{D}(\gamma), \beta)=\{\alpha,\beta\}
    \label{comp}
\end{equation}

In our model, $\mathfrak{C}$ is a simple operation achieved by simultaneously setting two or more values of the latent code  $c$ (that represents the meaning of the words) to extreme values as outlined in \citet{begus19,begusCiw}. Because the relationship between $\mathfrak{D}(\gamma)$ and $\beta$ exists, \ref{comp} is compositional.

While human languages have compositional properties, we have little evidence that neural processing of language is compositionally represented at the single neuron level in the human brain. The proposed model thus also  allows simulating single-neuron effects on compositional syntax and generating predictions that can be tested in the human brain.  The advantage of this model is that it operates with fully continuous data, which is a more realistic approximation of human syntactic processing---human syntax is processed from raw unsupervised spoken language. Other models operate exclusively at the symbolic level.

That inhibitory neurons play an important role in human language processing in the brain has been well established. The role of inhibitory neurons has primarily been established for segmentation \citep{giraud12}. Segmentation is closely related to concatenation, which lends some biological plausibility to our proposal.

This internal interpretability allows us to generate testable predictions for the evolution of one of the most consequential properties of human language: compositional syntax. That disinhibition might be a plausible candidate in the evolution and development of compositional concatenation might be suggested by the fact that disinhibition-associated subclasses of neurons (such as VIP and LAMP5) have been confirmed in the human middle temporal gyrus \citep{hodge19},  an area of the brain associated with the language network. Comparative studies also suggest that some of these classes of neurons are more common in primates compared to mice \citep{krienen20}. That syntactic processing develops at a relatively late stage and requires developments in the brain has been shown before \citep{hahne04,skeide14}. These assumptions are, however, highly speculative at this point and require further evaluation.

\subsection{The role of imitation}

While disinhibition is a likely explanation for the concatenated outputs in the negative latent space, it needs to be noted that concatenated outputs happen also when the latent code values are positive (Figure \ref{fig:numwordsbits}). 

The training data never provides evidence for concatenated outputs. It should be easy for the Discriminator to distinguish single- from multiple-word outputs and estimate multiple-word outputs as diverging from the real data distributions. One possible reason for why spontaneous concatenation is so common in GANs and why it occurs also in the positive values of the latent space is that unlike any other architecture, GANs learn primarily by imitation and they do not replicate data, but rather learn to construct data by innovation. In other words, in the GAN architecture, the agent that generates data (the Generator) never accesses training data directly, but learns to construct data from noise in the latent space by generating data such that another agent is unable to distinguish it from training data --- effectively forcing the imitative principle. Imitation is a prominent facilitator of language acquisition in humans, especially in early spoken language learning \citep{clark77,kuhl96,masur99,rasilo17}.

It is possible that the imitative and innovative design of GANs mimics the necessary conditions for the switch from simple to compound signals in humans. Learning by imitation and innovation is a crucial condition in human language. It is possible that human language learners start  uttering simple signals as concatenated and analyzing them as complex signals due to disinhibition as well as the imitative nature of learning. It has been shown elsewhere that the internal representations in ciwGAN/fiwGAN models closely match brain responses to sound of language in raw untransformed form \citep{begusZhouZhao23}, which makes the fiwGAN/ciwGAN models a plausible simulation of human speech processing in the brain.

\section{Conclusion}

Modeling the evolution from holistic single-call units into a compositional syntax is not a trivial task.  How human syntax emerged or how humans evolved from the stage of holistic calls into a fully concatenative and then compositional syntax is an unsolved problem. While other studies offer proposals about how this shift may have happened \citep{jackendoff99,luuk14,progovac15}, our paper shows that the shift spontaneously occurs in a computational model. Many studies exist that model evolution of syntax \citep{christiansen03}, but the present study is, to our knowledge, the first that shows how concatenation and compositionality emerge spontaneously in a fully connectionist model that did not have any explicit predispositions for concatenation and that is trained on raw speech data.

Our results suggest that the Generator network in the ciwGAN architecture not only learns to encode information that corresponds to lexical items in its audio outputs, but also spontaneously concatenates those lexical items into novel unobserved two-word or three-word sequences.

We also provide evidence that the concatenation is at least partly compositional. The latent code is compositional. When negative values representing a word get combined with positive values representing another word, the output is a predictable concatenation of the two words. Our approach thus proposes that a causal compositionality for a subset of codes and words emerges in a fully connectionist model trained on raw speech.

We propose a neural mechanism for the emergence of compositional concatenation based on \textit{disinhibition}. This explanation suggests that our modeling can be useful for generating testable biological and artificial neural predictions about language acquisition and evolution.  

The ability of unsupervised deep neural networks trained on raw speech to compositionally concatenate words into novel unobserved combinations has far-reaching consequences. This means that we can model basic syntactic properties directly from raw acoustic inputs of spoken language, which opens up potential to model several other syntactic properties directly from speech with ciwGAN/fiwGAN.%

From the perspective of evolution of syntax, the results suggest that a deep neural network architecture with no language-specific properties can spontaneously begin generating concatenated signals from simple signals. Precursors to compositionality emerge in a model that is trained on raw audio with no compositional training data points. In other words, concatenation and partial compositionality self-emerge in our models from raw audio based on single unit training data. The step from one-word stage to two-word stage is necessary both in evolution of human language as well as during language acquisition. We argue that unsupervised deep learning models not only concatenate single words into multi-word outputs, but are also able to embed words into novel unobserved combinations once the model is trained on multiple-word inputs.

Further research into the relationship between basic syntactic properties that spontaneously emerge in these fully unsupervised models trained on raw speech and the structure of the latent space has the potential to yield insights for the study of syntactic theory, language acquisition,  language evolution, and neuroscience. By evaluating these models on syntactic properties of spoken language, we should also get a better understanding of computational limits of unsupervised CNNs trained on raw speech.

Finally, modeling syntax from raw speech with deep neural networks is informative not only for cognitive science and linguistics, but for machine learning research in general as well. Speech processing increasingly by-passes text \citep{lakhotia-etal-2021-generative}. Understanding syntactic capabilities of spoken language models can provide information for future architectural choices as natural language processing expands from text to raw audio modeling.

\section*{Acknowledgements}

We would like to thank Bruno Ferenc \v{S}egedin for his useful comments.

\section*{Author contributions}

 First and corresponding author: Ga\v{s}per Begu\v{s}. 
 Spontaneous compositional concatenation and disinhibition were proposed by G.B. G.B.~and T.L. ran the experiments and wrote the paper. Z.W.~helped with initial paper plans. 

\section*{Data availability}
Generated audio files, code for statistical analysis, and models' checkpoints are available at this open source repository: \url{https://doi.org/10.17605/OSF.IO/PRZUQ}.

\section*{Conflict of interest}
The authors declare no conflicts of interest.

  \bibliographystyle{elsarticle-harv}

\newpage

\appendix
\section{Appendix}
\label{sec:appendix}

\begin{table}
 \begin{tabular}{r|l|r|l|r|l}
        \hline\hline
         \multicolumn{2}{c|}{Generator} & \multicolumn{2}{c|}{Discriminator}& \multicolumn{2}{c}{Q-Network}\\
         \hline
         Layer &  Dimension & Layer & Dimension & Layer & Dimension\\
         \hline
         $z$ & 100 $\times$ 1 & input & 32768 $\times$ 1 &input & 32768 $\times$ 1\\
         fc + reshape & 16 $\times$ 2048 & conv0 & 8192 $\times$ 64 &conv0 & 8192 $\times$ 64\\ 
        upconv0 & 64 $\times$ 1024 & conv1 & 2048 $\times$ 128 & conv1 & 2048 $\times$ 128\\ 
         upconv1 & 256 $\times$ 512 & conv2 &  512 $\times$ 256 & conv2 &  512 $\times$ 256\\ 
         upconv2 & 1024 $\times$ 256 & conv3 &  128 $\times$ 512 & conv3 &  128 $\times$ 512\\
        upconv3 & 4096 $\times$ 128 & conv4 & 32 $\times$ 1024 & conv4 & 32 $\times$ 1024\\
         upconv4 & 16384 $\times$ 64 & conv5 & 16 $\times$ 2048 & conv5 & 16 $\times$ 2048\\
         upconv5 & 32768 $\times$ 1 &flatten + logit & 1 $\times$ 1 &flatten + logit & 1 $\times$ 1 \\
         \hline\hline
    \end{tabular}
\caption{The structure of the Generator, the Discriminator, and the Q-network in the two-second experiments (based on \citealt{donahue19,begusCiw}). } \label{structure}
\end{table}

\begin{table}[ht]
\centering\scalebox{.7}{
\begin{tabular}{rrrrr}
  \hline
 & Estimate & Std. Error & z value & Pr($>$$|$z$|$) \\ 
  \hline
(Intercept) & -0.65 & 0.02 & -36.7 & 0.0000 \\ 
  bit1 & -0.21 & 0.00 & -45.3 & 0.0000 \\ 
  bit2 & -0.11 & 0.00 & -24.4 & 0.0000 \\ 
  bit3 & 0.01 & 0.00 & 1.6 & 0.1013 \\ 
  bit4 & -0.19 & 0.00 & -41.2 & 0.0000 \\ 
  bit5 & -0.17 & 0.00 & -39.8 & 0.0000 \\ 
  bit1:bit2 & 0.03 & 0.00 & 30.5 & 0.0000 \\ 
  bit1:bit3 & -0.02 & 0.00 & -14.7 & 0.0000 \\ 
  bit2:bit3 & -0.01 & 0.00 & -7.0 & 0.0000 \\ 
  bit1:bit4 & -0.00 & 0.00 & -1.7 & 0.0911 \\ 
  bit2:bit4 & 0.03 & 0.00 & 28.2 & 0.0000 \\ 
  bit3:bit4 & -0.02 & 0.00 & -17.8 & 0.0000 \\ 
  bit1:bit5 & 0.02 & 0.00 & 22.7 & 0.0000 \\ 
  bit2:bit5 & -0.03 & 0.00 & -29.7 & 0.0000 \\ 
  bit3:bit5 & -0.01 & 0.00 & -7.2 & 0.0000 \\ 
  bit4:bit5 & 0.04 & 0.00 & 33.2 & 0.0000 \\ 
  bit1:bit2:bit3 & -0.00 & 0.00 & -13.9 & 0.0000 \\ 
  bit1:bit2:bit4 & -0.01 & 0.00 & -32.4 & 0.0000 \\ 
  bit1:bit3:bit4 & 0.01 & 0.00 & 39.1 & 0.0000 \\ 
  bit2:bit3:bit4 & -0.00 & 0.00 & -9.1 & 0.0000 \\ 
  bit1:bit2:bit5 & -0.00 & 0.00 & -1.9 & 0.0551 \\ 
  bit1:bit3:bit5 & -0.00 & 0.00 & -16.0 & 0.0000 \\ 
  bit2:bit3:bit5 & 0.00 & 0.00 & 0.5 & 0.5923 \\ 
  bit1:bit4:bit5 & -0.00 & 0.00 & -12.4 & 0.0000 \\ 
  bit2:bit4:bit5 & -0.01 & 0.00 & -19.8 & 0.0000 \\ 
  bit3:bit4:bit5 & 0.00 & 0.00 & 2.7 & 0.0060 \\ 
  bit1:bit2:bit3:bit4 & 0.00 & 0.00 & 2.5 & 0.0135 \\ 
  bit1:bit2:bit3:bit5 & 0.00 & 0.00 & 15.1 & 0.0000 \\ 
  bit1:bit2:bit4:bit5 & 0.00 & 0.00 & 17.8 & 0.0000 \\ 
  bit1:bit3:bit4:bit5 & -0.00 & 0.00 & -6.8 & 0.0000 \\ 
  bit2:bit3:bit4:bit5 & 0.00 & 0.00 & 3.5 & 0.0005 \\ 
  bit1:bit2:bit3:bit4:bit5 & -0.00 & 0.00 & -9.8 & 0.0000 \\ 
   \hline
\end{tabular}}
\caption{Estimates of the logistic regression mixed effects model with the proportion of two-word outputs (one-word output = failure, two-word output = a success) as the dependent variable and individual bits with all interactions (including the five-way interaction) as the predictors. The random effect structure involves only the random intercept for each of the 10 unique random latent space samples $z$.}
\label{estimates}
\end{table}

 \begin{table}
 \centering
    \scalebox{.9}{\begin{tabular}{p{1in}cccc}
    & \multicolumn{4}{c}{Whisper annotated \# of words} \\
    \cmidrule{2-5}
        Human annotated \# of words & 1 &  2 & 3 &Total\\
    \midrule 
        1          &   47   &     9  & 1 & 57    \\     
        2           &   3   &     28  & 5   &36  \\
        3 & 0 & 2 & 5 & 7\\
            \midrule 

        Total & 50 & 39 & 11 & 100\\
    \bottomrule
    \end{tabular}}

    \caption{The accuracy of the Whisper speech-to-text model is assessed over 100 audio samples generated randomly using the two-second one-word ciwGAN model. Each sample was annotated manually by the authors and automatically using Whisper. For the classification of one word versus multiple words, Whisper agreed with the authors 88\% of the time.}
\label{whisper-accuracy}
 \end{table}

\begin{figure}
    \centering

        \includegraphics[width=.4\textwidth]{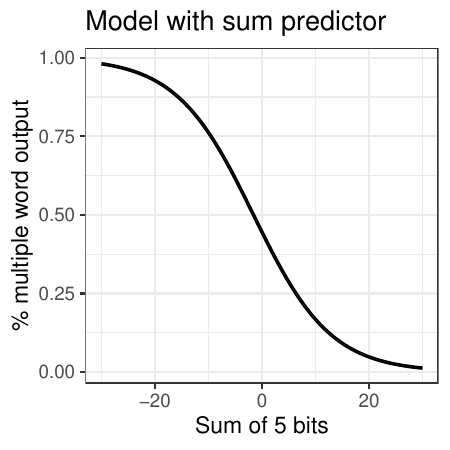}

    \caption{Predicted values of the logistic regression model with the proportion of two-word outputs (one-word output = failure, two-word output or more = a success) as the dependent variable and sum of bits 1--5 as the predictor with 95\% confidence limits (Model 4 in Table \ref{tab:parameters}). No random effects were included because mixed effects models result in singular fits. Outputs with no transcribed words were removed. The values were obtained with the \textit{effects} package \citep{effects1,effects2}. }
    \label{fig:var5single2srun600ggplot}
\end{figure}

\begin{figure}
    \centering

        \includegraphics[width=1\textwidth]{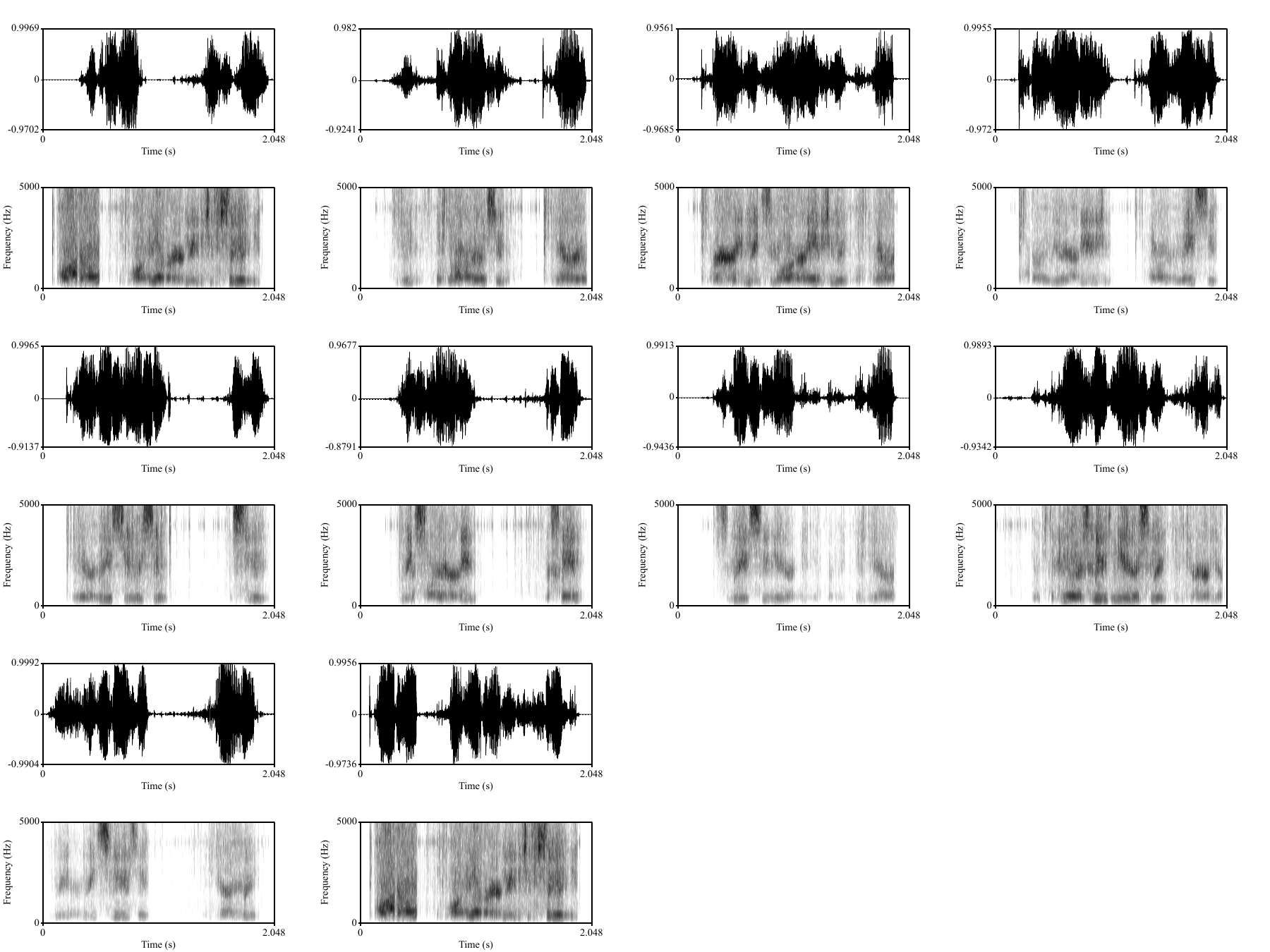}

    \caption{Ten randomly generated examples with the latent code [-1, -1, -1, -1, -1] of the single batch model.}
    \label{fig:oneBatch}
\end{figure}

\begin{figure}
    \centering
    \includegraphics[width=0.9\linewidth]{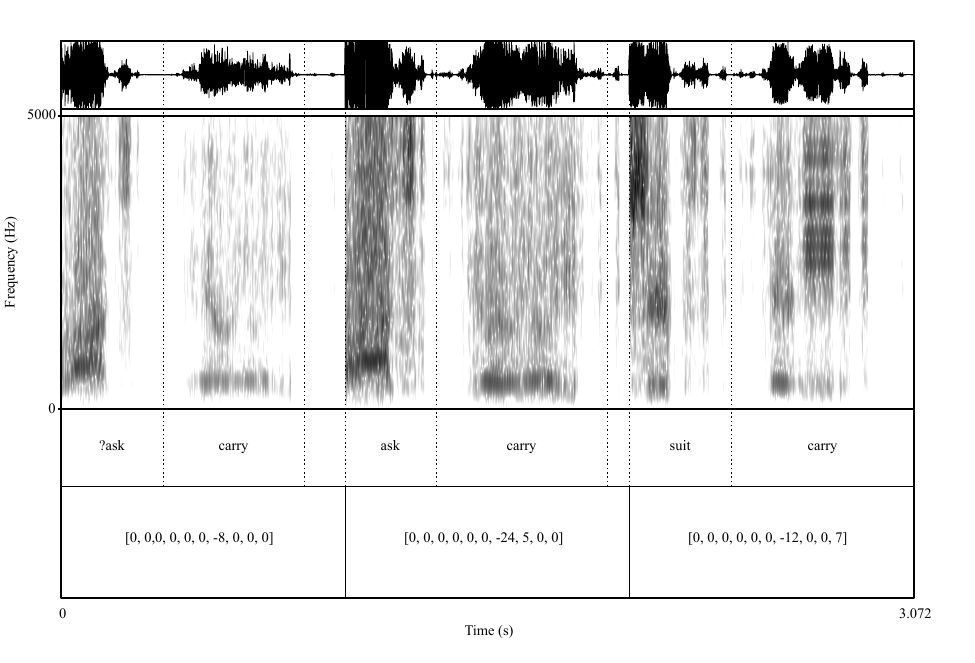}
    \caption{Waveforms and spectrograms (0-5 kHz) of three generated outputs with latent code variables as described in Table \ref{tab:codescomp}. The first output (from left) represent \textit{carry} with negative values of the latent code $c$. The word \textit{carry} is preceded by another word which is difficult to identify, but has some acoustic properties of \textit{ask}. The following outputs have compositional values of the latent code $c$: the negative value associated with \textit{carry} is combined with the positive value that represents individual lexical items. The outputs contain compositionally concatenated words with predictable word-pairs based on the latent code.}
    \label{fig:askCarry}
\end{figure}

\end{document}